\documentclass[10pt,twocolumn,letterpaper]{article}

\usepackage{cvpr}              

\usepackage{indentfirst} 

\usepackage{amsmath}
\usepackage{algorithm}
\usepackage{algorithmic}
\usepackage{balance}
\usepackage[accsupp]{axessibility}  
\usepackage{lineno}
\definecolor{cvprblue}{rgb}{0.21,0.49,0.74}
\usepackage[pagebackref,breaklinks,colorlinks,allcolors=cvprblue]{hyperref}

\def\paperID{9689} 
\def\confName{CVPR}
\def\confYear{2025}

\title{3D Dental Model Segmentation with Geometrical Boundary Preserving}

\author{
Shufan Xi$^{1}$, Zexian Liu$^{1}$, Junlin Chang$^{1,3}$, Hongyu Wu$^{1}$\thanks{Corresponding author.}, Xiaogang Wang$^{2}$, Aimin Hao$^{1}$ \\ 
$^1$State Key Laboratory of Virtual Reality Technology and Systems, Beihang University \\ 
$^2$College of Computer and Information Science, Southwest University \\ 
$^3$Peng Cheng Laboratory\\
{\tt\small \{xsfvrlab, liuzexian, changjunlin, whyvrlab, ham\}@buaa.edu.cn, wangxiaogang@swu.edu.cn}
}

\begin{document}
\maketitle
\begin{abstract}

3D intraoral scan mesh is widely used in digital dentistry diagnosis, segmenting 3D intraoral scan mesh is a critical preliminary task.
Numerous approaches have been devised for precise tooth segmentation. 
Currently, the deep learning-based methods are capable of the high accuracy segmentation of crown.
However, the segmentation accuracy at the junction between the crown and the gum is still below average.
Existing down-sampling methods are unable to effectively preserve the geometric details at the junction.
To address these problems, we propose CrossTooth, a boundary-preserving segmentation method that combines 3D mesh selective downsampling to retain more vertices at the tooth-gingiva area, along with cross-modal discriminative boundary features extracted from multi-view rendered images, enhancing the geometric representation of the segmentation network.
Using a point network as a backbone and incorporating image complementary features, CrossTooth significantly improves segmentation accuracy, as demonstrated by experiments on a public intraoral scan dataset. The source code is available at \url{https://github.com/XiShuFan/CrossTooth_CVPR2025}

\end{abstract}    
\section{Introduction}
\label{sec:introduction}

\begin{figure}[t]
  \centering
  \includegraphics[width=\linewidth]{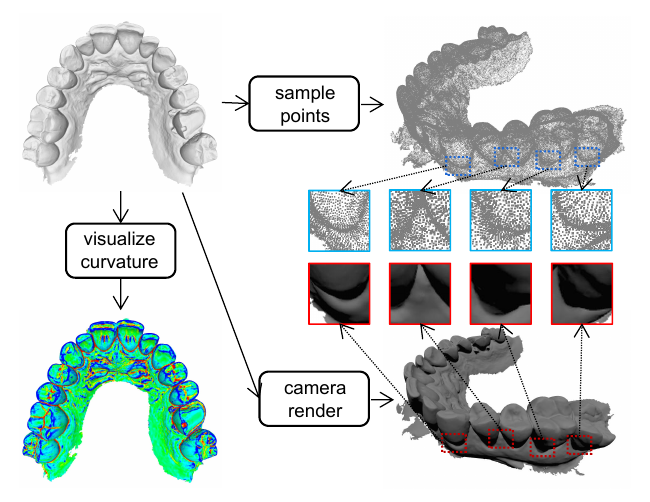} 
  \caption{An illustration of 3D intraoral scan model. The original intraoral scan consists of points and triangles, it can be visualized with a mean curvature histogram. The more red the color is, the lower the curvature is. Deep learning methods usually take sampled points as input, with blurry boundaries. But we can observe more clear tooth edges in images rendered from intraoral scan than in sampled points, indicated by red and blue boxes in the zoomed views respectively.}
  \label{fig:inspiration}
\end{figure}

Intraoral scanning is widely used in the field of digital dentistry and the crown segmentation is the key step. The main challenges \cite{20233dteethseg} in crown segmentation arise from the intrinsic similarity between crown shapes, the close arrangement of adjacent teeth, as well as abnormal teeth with irregular shapes. Additionally, external devices such as braces can influence the crown structure, making it difficult to crown segmentation.

To address these challenges, various methods have been proposed. Traditional methods rely on manually defined features for segmentation, including curvature-based methods \cite{wu2014tooth,yuan2010single}, contour-based methods \cite{sinthanayothin2008orthodontics}, and harmonic field-based \cite{liao2015automatic,zou2015interactive} methods. Frequently used curvature-based methods depend heavily on the intraoral scanner accuracy and the teeth condition of patients. In areas where curvatures are not apparent or in regions with steep curvatures like tooth crowns, these methods perform poorly and require manual correction, which lacks robustness.

Deep learning methods, driven by data, have surpassed traditional ones. Several works have applied deep learning methods directly on 3D intraoral scan models for segmentation. Some methods \cite{lian2020deep,tan2023boundary,tang2022contrastive,zhao20213d,zheng2022teethgnn} build networks in local-global feature fusion paradigm, where the network learns both local details and global context information from the point clouds. Some \cite{yuan2023full,zanjani2019deep} follow the encoder-decoder structure to integrate multi-scale features using skip connections and layer-wise aggregation to better percept small objects. Tooth centroid and bounding box detection methods \cite{cui2021tsegnet,qiu2022darch,tian20223d}, inspired by object detection networks like Faster-RCNN \cite{ren2016faster}, first detect every single tooth and then segment the cropped regions for higher resolution. Latest methods \cite{duan20233d,zhang2021tsgcnet,jana20233d} decouple coordinates and normal vectors from intraoral scan triangles, further improving performance by treating them as two different streams for learning.

Accurate identification of crown-gingiva boundary is critical to the subsequent data processing work. While existing methods have already achieved high segmentation accuracy, the boundary area between crown-gingiva is still not well treated.  This is because current deep learning based methods can not retain enough boundary details due to uniform mesh downsampling and fail to fully leverage the network's ability to learn boundary features merely based on coordinates and normal vectors, as shown in \cref{fig:inspiration}.

To address the above problems, we propose a boundary-preserving segmentation method named CrossTooth. We take curvature information from intraoral scans and dense features from rendered images into account, improving the overall segmentation performance, especially at tooth-gingiva areas. CrossTooth uses a carefully designed selective downsampling method to preserve triangles from tooth boundary while reducing the intraoral scan mesh to a specified number of triangles. Additionally, we extract discriminative boundary features from multi-view rendered images, which complement the sparse point cloud features.

The main contributions of our work can be summarized as follows:

\begin{itemize}[left=1em]
    \item We propose a selective downsampling method. According to the curvature of the intraoral scan model, this method can retain more geometric details and providing discriminative crown-gingival boundary features.
    
    \item We use multi-view rendering under specific lighting conditions to generate the distinguish shading on crown-gums boundary. By projecting rendered image features back to the point cloud, our method can enrich the sparse point cloud data with dense image features.
 
    \item Our CrossTooth is evaluated on a public intraoral scan dataset \cite{20233dteethseg}. The experimental results show that our CrossTooth significantly outperforms state-of-the-art segmentation methods.

\end{itemize}

\section{Related Work}
\label{sec:related work}

\begin{figure*}
  \centering
  \includegraphics[width=\textwidth]{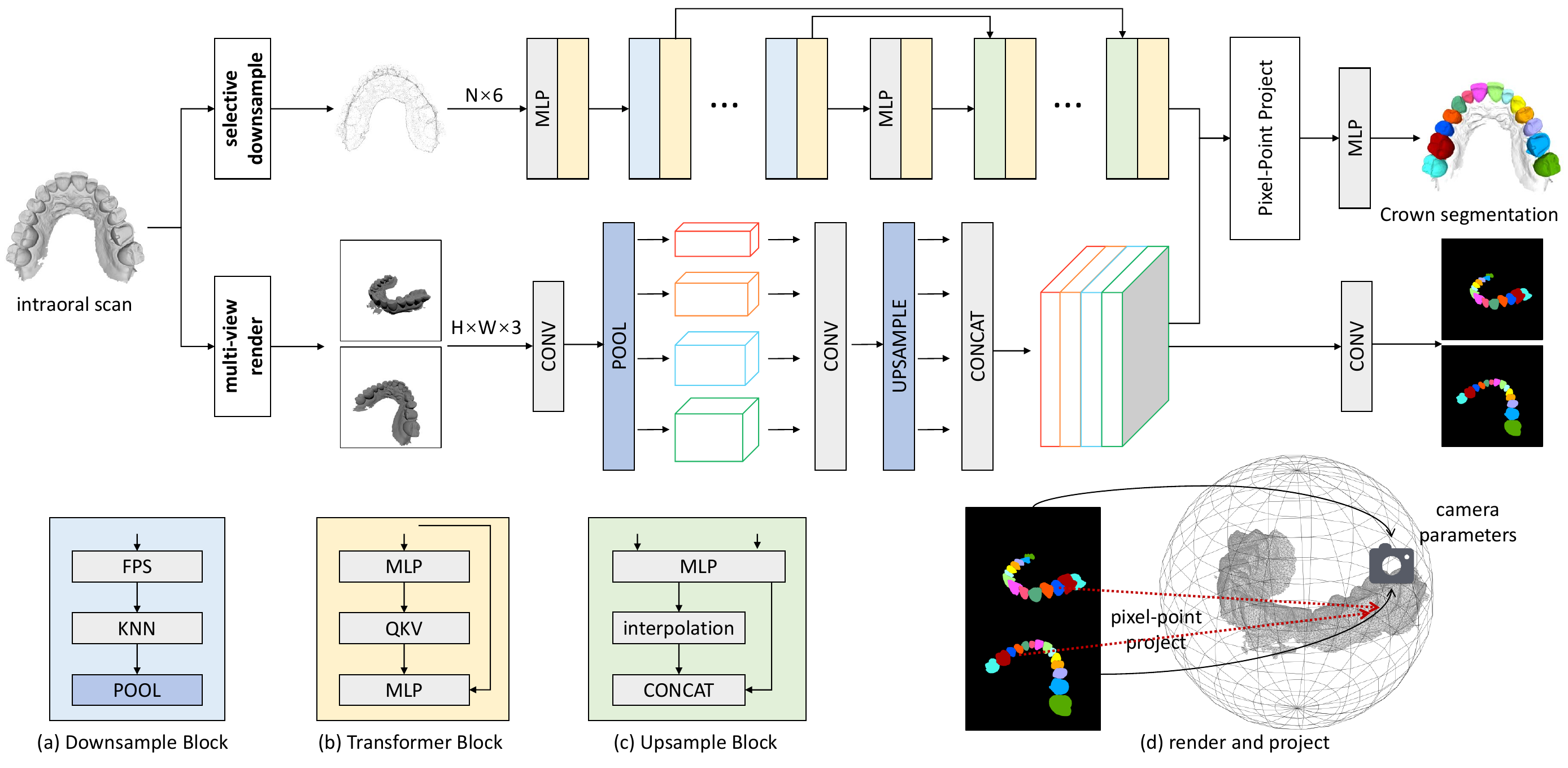} 
  \caption{Architecture of CrossTooth. The point network takes points from the intraoral scan model after selective downsampling as inputs and adopts a multi-scale encoder-decoder structure. The downsample block uses kNN to aggregate features from neighbor points, the transformer block applies a self-attention mechanism to learn long sequence contextual information, and the upsample block fuses features from the encoder and decoder, illustrated by (a) to (c) respectively. Image network takes rendered pictures and concatenates local-global features for downstream tasks. Then, following correspondences between the image and point, image features are projected back onto the point for further fusion illustrated by (d). Common MLP and CNN are used to produce final segmentation masks.}
  \label{fig:CrossTooth}
\end{figure*}

\subsection{Point Cloud Downsampling}

A point cloud is a form of 3D data representation composed of numerous unordered points, each characterized by spatial coordinates, color, and normal vectors. In deep learning tasks, the point cloud needs to be downsampled to a fixed number to form small batches. Popular downsampling methods include voxel-grid sampling, farthest point sampling, and quadric error metric sampling. 

Voxel grid sampling \cite{zanjani2019deep,tian2019automatic} divides the point cloud into small 3D grids and uses the center point of each grid to represent the original point set. Farthest point sampling \cite{qi2017pointnet} simplifies the point cloud by recursively selecting the farthest points from the current set, ensuring a uniform distribution of sampled points. Quadric error metric \cite{garland1997surface} sampling constructs a loss function to measure the geometric error introduced by merging vertices, ensuring both sampling quality and computational efficiency. Among the three methods, QEM is the most used, which is particularly effective in preserving edge and triangle topology.

However, in the intraoral scan segmentation task, all of these downsampling methods fail to retain the details at the boundaries. The loss of edge triangles leads to blurred boundaries and significantly impacts dense prediction tasks such as point cloud segmentation.

\subsection{3D Intraoral Scan Segmentation}

Point cloud segmentation methods are generally divided into voxel-based and point-based approaches, with the latter being more commonly used. PointNet \cite{qi2017pointnet}, as the pioneering point network, uses multi-layer perceptrons and pooling layers to learn global features. However, it could not perceive local relationships in point clouds, leading to the development of PointNet++ \cite{qi2017pointnet++}, which implements a hierarchical feature extraction and fusion strategy. DGCNN \cite{wang2019dynamic} further introduces the EdgeConv operator, learning contextual information through linear aggregation of central point features and edge features. Lastly, Point Transformer \cite{zhao2021point} provides a transformer-based framework for point networks, achieving state-of-the-art performance.

To accurately segment teeth from intraoral scan, Zhao \etal \cite{zhao20213d} constructs graph relationships among triangles, and uses graph attention convolution layers to extract fine-grained local geometric features and global features for comprehensive information. Lian \etal \cite{lian2020deep} proposes MeshSegNet based on PointNet, integrates adjacent matrices to extract multi-scale local contextual features. Xiong \etal \cite{xiong2023tsegformer} introduces TSegFormer with transformer layers, leveraging self-attention to capture dependencies between teeth while learning complex tooth shapes and structures.

The above methods generally treat global and local features in separate branches, leading to insufficient fusion between high- and low-level features. Other works have explored encoder-decoder methods with multi-scale feature fusion. Yuan \etal \cite{yuan2023full} proposes using a preliminary feature extraction module and weighted local feature aggregation module to retain fine details in tooth-gingiva boundaries. Zanjani \etal \cite{zanjani2019deep} introduces an end-to-end framework based on PointCNN \cite{li2018pointcnn}, employing Monte Carlo-based non-uniform resampling to train and deploy models at the highest available spatial resolution. 

Recently, some studies have found that separating triangle coordinates and normal vectors into two branches improves performance. Zhang \etal \cite{zhang2021tsgcnet} introduces TSGCNet, a dual-branch graph convolutional network that learns multi-view geometric information from both coordinates and normal vectors. Duan \etal \cite{duan20233d} extends this strategy with SGTNet, using a graph transformer module to enhance boundary classification accuracy.


\subsection{Point Cloud Edge Segmentation}

The accuracy of point cloud boundary segmentation is crucial, especially in indoor scene segmentation, outdoor road segmentation, and medical organ segmentation tasks.

Hu \etal \cite{hu2020jsenet} proposes a dual-branch fully convolutional network on S3DIS and ScanNet datasets for joint semantic segmentation and edge detection. With a boundary map generation module and a well-designed edge loss function, the network is constrained to achieve more precise segmentation at object boundaries. Tang \etal \cite{tang2022contrastive} analyzed the performance of existing point cloud segmentation methods at scene boundaries, and proposed the Contrastive Boundary Learning (CBL) framework to optimize boundary distinctiveness. CBL contrasts boundary point features during point cloud downsampling stages by push and pulls operations on boundary points from different classes. Gong \etal \cite{gong2021boundary} introduces a Boundary Prediction Module to predict boundaries by discriminative aggregation of features within a neighborhood, preventing local features of different categories from interfering with each other. Liu \etal \cite{liu2023grab} is the first to focus on boundary segmentation in medical point cloud tasks, proposing a graph-based boundary-aware network for intracranial aneurysm and tooth point cloud segmentation.

These above works demonstrate the growing importance of precise boundary segmentation in point cloud tasks, especially for complex scenarios.

\section{Methods}
\label{sec:methods}

\subsection{Overview}

We propose CrossTooth, a boundary-preserving method for intraoral scan segmentation illustrated in \cref{fig:CrossTooth}. Instead of relying on the commonly used QEM \cite{garland1997surface}, we develop a selective downsampling method, which better preserves boundary details. Additionally, multi-view image rendering is used to capture dense information, with an image segmentation network trained to learn discriminative boundary features. Finally, image features are projected back onto the point cloud for a more comprehensive understanding.

Different from existing methods, our CrossTooth adopts a complementary approach by learning feature representations from multi-view rendered images and discrete point clouds respectively. By merging these two streams through MLP at the final stage, CrossTooth compensates for the information loss inherent in point clouds, as illustrated in \cref{fig:inspiration}, leading to improved dense segmentation predictions.

\subsection{Selective Downsampling}

The high-resolution intraoral scan model contains more than 100,000 points. Most works downsample intraoral scan to 16,000 points for training. Frequently used downsampling methods fall into two categories. The first type uses point cloud downsampling methods \cite{zhang2022implicit,wang2022tooth,li2020malocclusion,cui2021tsegnet,liu2023grab,qiu2022darch,liu2022hierarchical,zhuang2023robust}, which cannot preserve the structure of edges and triangles. The second type performs surface reduction processing \cite{zhang2021tsgcnet,chen2023automatic,krenmayr2024dilatedtoothsegnet,li2024fine,wu2022two,xu20183d,duan20233d,jana20233d,li2022multi,lian2020deep,sun2020automatic,sun2020tooth,zhao20213d,zheng2022teethgnn} on the model, which can maintain the original topological structure of the model. However, even the second type cannot retain tooth boundary information. 

We propose a novel boundary-preserving downsampling method for intraoral scan models, termed selective downsampling. This approach builds upon the well-established QEM \cite{garland1997surface} downsampling method and incorporates curvature prior information to enhance its performance.
Here, we provide a brief review of QEM: 


The QEM algorithm simplifies a model by merging two edge points  \( v_1 \) and \( v_2 \) into a single one \( v \), while keeping \( v \) as close as possible to the original model. The square of the Euclidean distance between \( v \) and the corresponding local surface of the model is used as the error metric. Define \( \text{plane}(v) \) as the original triangles corresponding to point \( v \), the optimization function is:

\begin{equation}
    v={arg min}_v\sum_{\substack{p \in plane(v_1) \\ \cup plane(v_2)}}k *distance(v,p)^2,
\end{equation}
where \( k \) is the weight of the distance from point \( v \) to each related triangles. This optimal solution of target \( v \) can be achieved by solving the extremum of a quadratic function.

In the original QEM algorithm, \( k \) is set to 1 for the same weight, whereas in our selective downsampling algorithm, we set \( k \) to be a value related to the curvature of points \( v_1 \) and \( v_2 \). Curvature describes the degree of bending of a geometric surface, noticeable negative curvature can be observed at tooth boundaries, while sharp positive curvature can be seen at the top of tooth crowns. We multiply a larger coefficient for negative curvature edges (\( k = 10 \)) and a smaller coefficient for positive curvature edges (\( k = 1 \)) during the QEM iterations. After each round of iteration, the curvature of the current model is recalculated. The algorithm for this process is shown in \cref{alg:Selective Downsample}.

\begin{figure}[t]
  \centering
  \includegraphics[width=\linewidth]{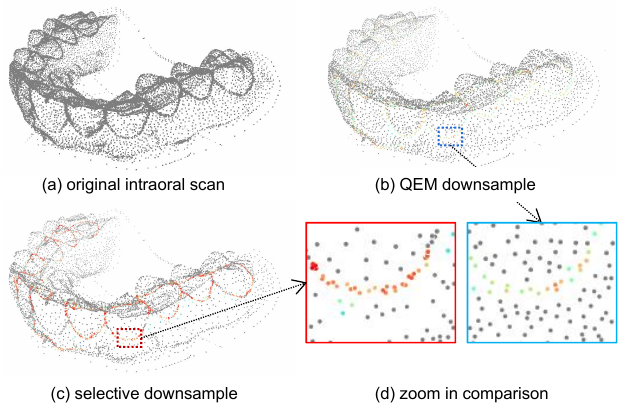} 
  \caption{Comparison of QEM and selective downsampling method. Our method performs better than QEM in tooth boundaries, as visualized in density histogram (d). Quantitatively, selective downsampling results in 10\% to 15\% density more than QEM at boundary areas.}
  \label{fig:sd}
\end{figure}

The selective downsampling algorithm fully considers the curvature of the intraoral scan model, it applies mild and aggressive downsamples in tooth boundary and other areas respectively. The comparison of QEM and selective downsampling is illustrated by \cref{fig:sd}. We also give a quantitative metric to show how well our method preserves boundary points as shown in \cref{tab:density}.

\begin{algorithm}
    \caption{Selective Downsampling}
    \label{alg:Selective Downsample}
    \renewcommand{\algorithmicrequire}{\textbf{Require:}}
    \renewcommand{\algorithmicensure}{\textbf{Ensure:}}
    \begin{algorithmic}[1]
        \REQUIRE A mesh $M$
        \ENSURE A simplified mesh $M$
        
        \STATE Compute the $Q$ matrices for all the initial vertices.
        \STATE Compute mean curvature $H$ for all the initial vertices.
        \STATE Select all valid pairs.
        \STATE Compute the optimal contraction target $\bar{v}$ for each valid pair $(v_1, v_2)$. The error $\bar{v}^T(Q_1 + Q_2)\bar{v}$ of this target vertex $\bar{v}$ becomes the cost of contracting that pair.
        \STATE Compute the edge collapse coefficient by averaging the curvature of its two points. Multiply the cost by coefficient.
        \STATE Place all the pairs in a heap keyed on cost with the minimum cost pair at the top.
        \REPEAT
            \STATE Remove the pair $(v_1, v_2)$ of the least cost from the heap, contract this pair, and update the costs of all valid pairs involving $\bar{v}$
        \UNTIL the heap is empty
        \RETURN $M$
    \end{algorithmic}
\end{algorithm}

\begin{table}
  \centering
  \setlength\abovedisplayskip{3pt}
  \setlength\belowdisplayskip{3pt}
  \begin{tabular}{@{}lcc@{}}
    \toprule
    Method & Upper jaw \(\downarrow\) & Lower jaw \(\downarrow\) \\
    \midrule
    QEM & 5.655e-3 & 4.797e-3 \\
    Selective downsample & 5.029e-3 & 4.052e-3 \\
    \bottomrule
  \end{tabular}
  \caption{Average distance of tooth boundary points at the upper and lower jaw. We define boundary points following the criterion in \cite{liu2023grab} and set neighbor points to 4. It can be seen that the density of boundary points obtained by our method is 15\% higher.}
  \label{tab:density}
\end{table}

\subsection{Multi-View Image Rendering}

Images and point clouds are commonly utilized in multimodal learning \cite{yan20222dpass,chen2023cross}. Point clouds offer detailed geometric information,  but the geometric details is readily eliminated during the down-sampling process. Conversely, edge information in images tends to be better preserved during down-sampling process.

To enhance the presentation of boundary information in rendered images, we conducted tests with various parallel lights from different angles. We found that a vertical downward parallel light positioned above the crown yielded the most striking contrast at the gum-to-crown junction, which is shown in \cref{fig:inspiration}. To capture all such junctions comprehensively, we configured multiple perspective cameras on the upper hemisphere of the intraoral scan models for multi-view rendering. 
Specifically, the longitude and latitude of the upper hemisphere are evenly divided, the virtual cameras are placed on the sampling positions, pointing to the center of intraoral scan. The optical axis of virtual camera must pass through the center of intraoral scan model.

Point cloud coordinates can be mapped to 2D pixel coordinates with camera parameters. Once the mapping relationship is established, the corresponding image features can be fused into the point cloud, achieving modal complementarity. The advantages of performing multi-view rendering include comprehensive modeling of the intraoral scan model by capturing images from different angles, supplementing missing tooth boundary information in the point cloud, and reducing data bias while improving model robustness. 

\begin{figure*}
  \vspace{-10pt}
  \centering
  \includegraphics[width=\textwidth]{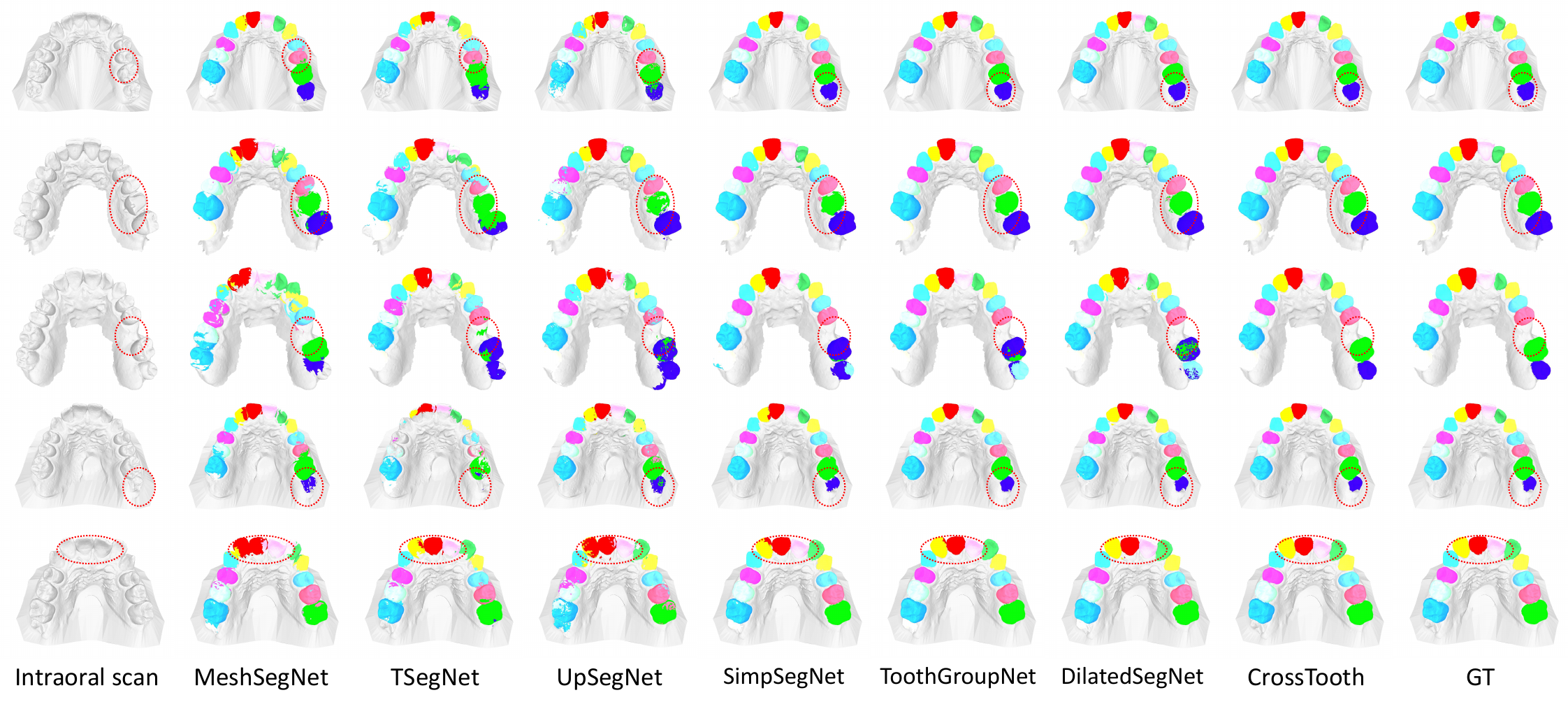} 
  \caption{Visualization of segmentation results, along with respective ground-truth annotations. Important areas are marked with red dotted circles. Our CrossTooth performs better than other methods under all the listed intraoral scan cases.}
  \label{fig:comparison}
  \vspace{-3pt}
\end{figure*}

\subsection{Boundary Aware Segmentation}

CrossTooth consists of two parts, the image segmentation module and the point segmentation module. The image module uses PSPNet \cite{zhao2017pyramid}, taking the rendered image as input and producing semantic segmentation results (C × H × W = 17 × 1024 × 1024). The point module takes the features of the intraoral scan model after selective downsampling as input. Each point's features include normalized spatial coordinates and spatial normal vectors, concatenated as a 6-dimensional vector. Specifically, we use Point Transformer \cite{zhao2021point}, which includes multi-stage downsampling encoders and upsampling decoders. In each stage, transformer layers are used for long-sequence contextual self-attention learning. In each stage, we use kNN to fuse neighbor point features. Finally, several layers of MLP is used as the segmentation head to output a multi-class segmentation mask. We also use another MLP to predict a binary segmentation mask for the tooth-gingiva boundary.

Considering that the image network and point network learn distinctly different feature representations from two complementary perspectives, merging their outputs allows the entire network to understand the structure of the intraoral scan model comprehensively. We concatenate the weighted image segmentation results with the point features from the last decoding layer of the point network as shown in \cref{fig:CrossTooth}. The pixel features \( F_{pixel} \) of each image viewpoint correspond to the point features \( F_{point} \) in the point cloud. We encode pixel features to one-hot vectors and concatenate them with the point features, followed by using a standard MLP for feature fusion, which can be represented as follows:
\begin{equation}
    F_{fusion}=\operatorname{MLP}(F_{point} \oplus encode(avg(F_{pixel}))).
\end{equation}

To explicitly constrain tooth boundaries, we follow previous works \cite{liu2023grab,tang2022contrastive} to perform supervised learning for tooth boundary segmentation. For a point, if more than half of its \( k = 8 \) nearest neighbors belong to different classes, the point is defined as a boundary point. Since point cloud networks involve multi-stage downsampling, making it difficult to distinguish boundaries in deep stages, we only compute the loss after the last decoder layer. In both the image and point cloud segmentation tasks, we use Cross Entropy Loss, as represented below:
\begin{equation}
    L_{CE}=-\sum_{i=1}^{N}\sum_{c=1}^{C}p_{ic} \log y_{ic},
\end{equation}
where \( p_{ic} \) and \( y_{ic} \) represent the predicted class probability and the true class probability for a pixel/point. \( N \) represents the total number of pixel/point, and \(C\) represents the total classes of teeth, which is fixed as 17 in our task.

In addition to explicitly predicting boundary points, we prefer the features learned by the point network to be close for neighboring points from the same class and far apart for neighboring points from different classes. We apply the CBL \cite{tang2022contrastive} loss function only on the last decoder layer, as represented below:
\begin{equation}
    L_{CBL}=-\frac{1}{|P|}\sum\limits_{x \in P}log{\frac{\sum\limits_{\substack{y \in N_x \\ L_x=L_y}} exp(-d(F_x,F_y))}{\sum \limits_{y \in N_x} exp(-d(F_x, F_y))}},
\end{equation}
where \( P \) is the set of all points in the point cloud, \( N_x \) represents the set of neighboring points within a certain radius for point \( x \), \( L_x \) represents the label of point \( x \), \( F_x \) represents the feature vector of point \( x \), and \( d(\cdot,\cdot) \) is the euclidean distance function. CBL loss encourages features to keep away from different classes, this guides the network to learn distinctive boundary features.

The overall loss function can be formulated as follows:
\begin{equation}
    Loss=L_{CE}(image,point)+L_{CBL}(point).
\end{equation}

Our segmentation process focuses on preserving boundary information. The point cloud is first processed through selective downsampling, which results in higher precision in the tooth boundary regions. Additionally, we extract multi-view rendered image features, particularly boundary details, and project them back onto the original point cloud for feature fusion. Experimental results demonstrate that our approach outperforms the state-of-the-art methods.

\section{Experiments}
\label{sec:experiments}

\begin{table*}[t]
  \centering
  \scalebox{0.9}{
  \begin{tabular}{@{}lcccccccccc@{}}
    \toprule
    Method & mIoU & Boundary & Background & T1/T9 & T2/T10 & T3/T11 & T4/T12 & T5/T13 & T6/T14 & T7/T15 \\
    \midrule
    MeshSegNet \cite{lian2020deep} & 66.130 & 0.900 & 86.812 & 62.779 & 49.343 & 56.577 & 60.870 & 58.972 & 66.249 & 40.132 \\
    SimpSegNet \cite{jana20233d} & 88.450 & 74.703 & 94.566 & 84.340 & 85.817 & 85.025 & 89.363 & 85.626 & 86.234 & 59.953 \\
    TSegNet \cite{cui2021tsegnet} & 57.239 & 0.000 & 80.694 & 53.274 & 50.026 & 50.103 & 55.560 & 49.544 & 59.189 & 27.043 \\
    TeethGNN \cite{zheng2022teethgnn} & 74.631 & 76.015 & 93.978 & 66.285 & 64.297 & 68.889 & 69.496 & 60.497 & 68.123 & 43.760 \\
    ToothGroupNet \cite{20233dteethseg} & 93.546 & 76.392 & 95.584 & 93.397 & 92.681 & 89.608 & 92.399 & 90.561 & 92.538 & 65.129 \\
    UpToothSeg \cite{he2022unsupervised} & 83.948 & 62.880 & 93.358 & 84.876 & 82.859 & 82.650 & 86.113 & 77.493 & 77.375 & 48.312 \\
    DilatedSegNet \cite{krenmayr2024dilatedtoothsegnet} & 91.441 & 74.899 & 95.558 & 92.562 & 92.089 & 89.001 & 91.766 & 88.521 & 89.229 & 62.699 \\
    \textbf{Ours} & \textbf{95.860} & \textbf{82.058} & \textbf{96.410} & \textbf{95.005} & \textbf{94.784} & \textbf{91.592} & \textbf{94.547} & \textbf{93.309} & \textbf{95.088} & \textbf{68.055} \\
    \bottomrule
  \end{tabular}
  }
  \caption{The segmentation results for seven competing methods and our CrossTooth. For brevity, we combine the metric of T1/T9 to T7/T15, as they are pairs of teeth with similar shapes, distributed symmetrically on the left and right sides of the mouth. We ignore T8/T16 as wisdom teeth are rare in our dataset.}
  \label{tab:comparison}
\end{table*}
\subsection{Data Preparation}

We use a dataset publicly available in MICCAI Challenge \cite{20233dteethseg} which includes a total of 1800 upper and lower jaw intraoral scan models. The dataset is randomly divided into 1440 for the training set and 360 for the testing set. Some dental models contain non-manifold edges, which we remove by splitting vertices in the pre-processing phase. To be consistent with previous works \cite{zhang2021tsgcnet,zanjani2019deep,lian2020deep,20233dteethseg}, we downsample the intraoral scan models to 16,000 points using selective downsampling. Data augmentation operations are performed online during the training phase. Specifically, after normalizing the coordinates, we apply random translations with a uniform distribution of \( [-0.1, 0.1]\) in the length, width, and height of its bounding box. Rotation is applied using angles generated from a normal distribution $N(0, 1)$. For each intraoral scan model, we render 96 images from different viewpoints to ensure most points are visible as shown in \cref{fig:CrossTooth}. In our rendering setup, a dental arch is firstly aligned using PCA analysis, then we employ a directional light source in Pyrender to simulate parallel light rays with pure white color. The intensity is specified as 2, providing enhanced brightness across the scene, which facilitates the visualization of details and improves clarity.

Our task is to automatically segment each model into 17 different semantic parts, including 16 different teeth (wisdom teeth also included) and the gingiva/background. To standardize the label of upper and lower teeth, we fixed the orientation of jaws and labeled each tooth from T1 to T16, with the gingiva labeled as T0, illustrated in \cref{fig:comparison}.

\subsection{Experiment Setting}

Our CrossTooth consists of an image segmentation module and a point cloud segmentation module. The image segmentation module uses the highly efficient PSPNet \cite{zhao2017pyramid}. The point cloud segmentation module is based on Point Transformer \cite{zhao2021point}, consisting of five encoding and decoding stages. In each encoding stage, the point cloud is progressively downsampled using FPS, and the number of points is gradually restored in each decoding stage. Channels for these five stages are 32, 64, 128, 256, and 512, with downsampling rates of 1, 4, 4, 4, and 4, respectively. The number of neighbor points used for kNN-based feature fusion in each stage is 8, 16, 16, 16, and 16. Throughout the network, we apply a combination of Batch Normalization and ReLU for non-linear transformations.

We also select seven existing intraoral scan segmentation methods for comparison, each named as MeshSegNet \cite{lian2020deep}, SimpSegNet \cite{jana20233d}, TSegNet \cite{cui2021tsegnet}, TeethGNN \cite{zheng2022teethgnn}, ToothGroupNet \cite{20233dteethseg}, UpToothSeg \cite{he2022unsupervised}, DilatedSegNet \cite{krenmayr2024dilatedtoothsegnet}. We use their official implementations and ensure they are under the same training conditions as with CrossTooth.

The input for all networks is an N × C vector, where N=16,000 represents the number of points, and C=6 represents the 3D coordinates and normal vectors for each point. All models are trained for 100 epochs using cross-entropy loss on an NVIDIA RTX 3090 GPU. We use the Adam optimizer with a mini-batch size of 4. The initial learning rate is set to 1e-3, and a cosine scheduler is employed for learning rate decay, with a minimum learning rate of 1e-6. We evaluate performance using the Intersection over Union metric for each tooth category, as well as the overall segmentation mIoU. Additionally, we test the tooth boundary segmentation IoU to demonstrate the superiority of our method.

\subsection{Tooth Segmentation}

The detailed segmentation results are shown in \cref{tab:comparison}. It demonstrates that our method achieves the best performance in both tooth segmentation IoU and tooth boundary segmentation IoU metrics. Specifically, compared to the best method in this task, ToothGroupNet, our CrossTooth improves tooth IoU and boundary IoU by 2.3\% and 5.7\%, respectively. Additionally, CrossTooth significantly outperforms other methods, further validating the effectiveness of our boundary preservation mechanism, which can learn more discriminative boundary features and achieve precise tooth segmentation.

We select several intraoral scan models and visualize the segmentation results of different methods, as shown in \cref{fig:comparison}. Visually, our CrossTooth outperforms the competing methods, especially in the challenging areas marked with red dotted circles in the figure. Specifically, MeshSegNet, TSegNet, and UpToothSeg reveal insufficient differentiation between adjacent teeth, leading to confusion between neighboring tooth triangles. SimpSegNet, ToothGroupNet, and DilatedSegNet show better segmentation performance but still produce isolated boundary segmentation noises. In contrast, CrossTooth takes advantage of complementary image information and learns more discriminative features for tooth boundaries, performing well across different oral conditions. In row 4, CrossTooth accurately segments the atrophied teeth at the end of the dental arch. Moreover, in row 3, where a tooth is missing from the middle of the dental arch, only CrossTooth correctly labels the remained teeth.

\subsection{Ablation Study}

\begin{figure}[t]
  \centering
  \includegraphics[width=\linewidth]{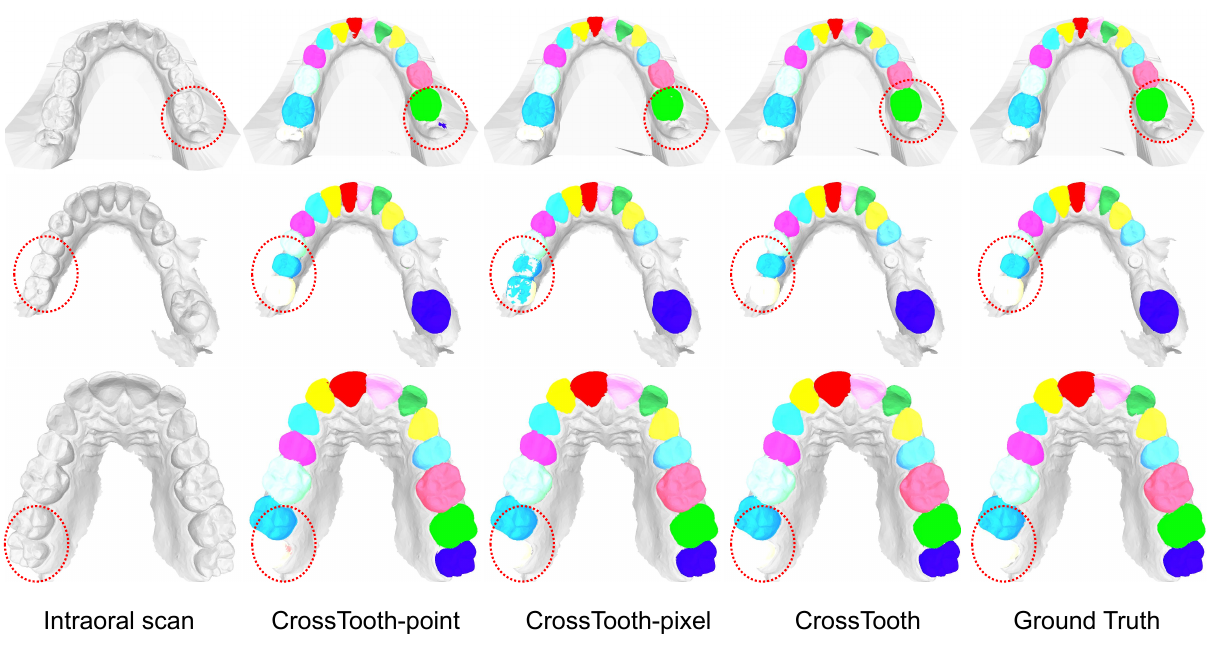} 
  \caption{CrossTooth performs better than the other two methods using only image or point features. It demonstrates that image and point features are complementary, they can eliminate each other's wrong segmentation results.}
  \label{fig:ablation}
\end{figure}

The fusion of dense image features, particularly tooth boundary features, contributed to the optimal performance of our method. In this section, we investigate the impact of integrating image features. Specifically, we trained CrossTooth-point, which does not incorporate image information, and CrossTooth-pixel, which uses only image information. The segmentation results of these three models under the same training conditions are shown in \cref{tab:ablation}. With the rich features from images, CrossTooth improves tooth segmentation IoU and tooth boundary segmentation IoU by 0.7\% and 0.5\% respectively. This demonstrates the complementarity between the geometric information provided by the point cloud and the texture details provided by the images. As shown in the circles in \cref{fig:ablation}, the mis-segmented areas by CrossTooth-point can be corrected through the fusion of point and image features, proving the effectiveness of our method.
\begin{table}[ht]
  \centering
  \begin{tabular}{@{}lcc@{}}
    \toprule
    Method & mIoU & Boundary IoU \\
    \midrule
    CrossTooth-point & 95.119 & 81.572 \\
    CrossTooth-pixel & 89.488 & - \\
    CrossTooth & 95.860 & 82.058\\
    \bottomrule
  \end{tabular}
  \caption{Segmentation metrics with \& w/o pixel or point feature. By integrating image and point features, the origin CrossTooth achieves the best performance.}
  \label{tab:ablation}
\end{table}

We further introduce another dataset named 3D-IOSSeg \cite{li2024fine} and two baselines TSGCNet \cite{zhang2021tsgcnet} and HiCA \cite{li2024novel}. Ablation studies on image numbers and selective downsampling are conducted, as shown in \cref{tab:image_ablation} and \cref{tab:downsample_ablation} respectively. The rendered images consistently improve performance across all methods. However, the minor improvement brought by image features may be due to the simple fusion strategy, which involves only a few MLPs. Given an existing method, selective downsampling can improve its performance, as shown in \cref{tab:downsample_ablation}. And CrossTooth outperforms the others even without selective downsampling.

\begin{table}[ht]
  \setlength{\abovecaptionskip}{5pt}
  \centering
  \begin{tabular}{@{}lccc@{}}
    \toprule
    \small{Method\textsubscript{FLOPs}} & \small{mIoU} & \small{Boundary IoU} \\
    \midrule
    \small{SimpSegNet\textsubscript{64.46G}} & \scriptsize{67.71/87.07/88.33/}\small{\textbf{88.46}} & \scriptsize{51.10/}\small{\textbf{66.99}}\scriptsize{/63.36/63.05} \\
    \small{ToothGroup\textsubscript{8.53G}} & \scriptsize{84.62/89.64/87.46/}\small{\textbf{89.76}} & \scriptsize{63.57/67.62/68.09/}\small{\textbf{68.44}} \\
    \small{DilatedNet\textsubscript{139.20G}} & \scriptsize{83.60/90.64/90.51/}\small{\textbf{90.74}} & \scriptsize{37.12/47.07/}\small{\textbf{53.55}}\scriptsize{/50.87} \\
    \small{TSGCNet\textsubscript{174.85G}} & \scriptsize{76.45/84.75/85.63/}\small{\textbf{85.94}} & \scriptsize{59.92/64.09/}\small{\textbf{65.30}}\scriptsize{/64.82} \\
    \small{HiCANet\textsubscript{97.11G}} & \scriptsize{78.77/86.26/87.92/}\small{\textbf{88.05}} & \scriptsize{64.78/66.01/}\small{\textbf{67.17}}\scriptsize{/66.57} \\
    \small{CrossTooth\textsubscript{5.05G}} & \scriptsize{86.11/87.88/88.59/}\small{\textbf{88.79}} & \scriptsize{65.30/66.21/}\small{\textbf{68.03}}\scriptsize{/66.65} \\
    \bottomrule
  \end{tabular}
  \caption{Ablation study on the number of rendered images in the 3D-IOSSeg dataset. Each method is evaluated under four conditions: no image, 32 images, 96 images, and 128 images. The segmentation mIoU improves as the number of images increases. The boundary IoU declines after a certain number of images. The PSPNet used in CrossTooth as an image feature extractor contains only 7.08G FLOPs, making it lightweight and not demanding in terms of computational resources.}
  \label{tab:image_ablation}
\end{table}

\begin{table}[ht]
  \setlength{\abovecaptionskip}{5pt}
  \centering
  \begin{tabular}{@{}lccc@{}}
    \toprule
    \small{Method} & \small{mIoU with \& w/o SD} & \small{B-IoU with \& w/o SD} \\
    \midrule
    \small{TSGCNet\textsuperscript{1}} & \textbf{91.22}/89.86 & \textbf{73.40}/70.33 \\
    \small{HiCANet\textsuperscript{1}} & \textbf{91.47}/90.15 & \textbf{75.48}/72.67\\
    \small{CrossTooth\textsuperscript{1}} & \textbf{95.86}/93.88 & \textbf{82.05}/80.73\\
    \midrule
    \small{TSGCNet\textsuperscript{2}} & \textbf{76.45}/75.02 & \textbf{59.92}/52.49 \\
    \small{HiCANet\textsuperscript{2}} & \textbf{78.77}/76.43 & \textbf{64.78}/61.28 \\
    \small{CrossTooth\textsuperscript{2}} & \textbf{86.11}/85.10 & \textbf{65.30}/62.07\\
    \bottomrule
  \end{tabular}
  \caption{Ablation study on selective downsampling. Superscript 1 refers to the 3DTeethSeg'22 dataset, while superscript 2 refers to the 3D-IOSSeg dataset. Each experimental group measures both mIoU and boundary IoU with and without selective downsampling. Selective downsampling can improve segmentation performance, especially in boundary areas.}
  \label{tab:downsample_ablation}
\end{table}
\section{Discussion}
\label{sec:discussion}

Although our CrossTooth achieves leading performance, it still has certain limitations. For example, CrossTooth does not handle cases with few teeth, in which scenario our method produces incorrect predictions on the tooth boundaries, this may be due to the tooth as a whole not being closely related to the tooth boundaries. We will include more cases with severe tooth loss as training samples in future studies and design robust post-processing steps. Besides, the varying number and appearance of wisdom teeth on each intraoral scan lead to most of the wrong predictions. It is hard for network to learn features of wisdom teeth with only few samples. We have taken wisdom teeth into consideration in our experiments but got low accuracy. So in future works we are to design few-shot learning strategy for wisdom teeth. Another shortage is the feature fusion layer, we only leverage simple MLP to extract complementary features at the last stage, which may be not fine-grained enough at local fields. We will consider more strategies such as multi-level encoder-decoder structure.
\section{Conclusion}
\label{sec:conclusion}

We propose CrossTooth to automatically segment individual teeth from intraoral scan models. To avoid existing downsampling methods from losing boundary details, we propose selective down-sampling based on curvature prior information. We fuse image dense features to improve the point cloud segmentation accuracy, especially retaining the detailed information at the tooth boundary. Compared with popular QEM downsample method, selective downsampling increases vertex density by 10\% to 15\%. By training on the public intraoral scan model dataset, it is proved that our method is superior to other competing methods, achieving 95.86\% and 82.05\% at overall mIoU and boundary IoU respectively. Our CrossTooth can effectively assist doctors in clinical diagnosis and analysis.

\noindent\textbf{Acknowledgment.} This work was supported in part by the National Natural Science Foundation of China under Grant 62132021, the Guangxi Science and Technology Major Program under Grant GuiKeAA24206017 and the National Key R\&D Program of China under Grant 2023YFC3604505.
{
    \small
    \bibliographystyle{ieeenat_fullname}
    \bibliography{main}
}


\end{document}